\documentclass[10pt,twocolumn,letterpaper]{article}

\usepackage{cvpr}              

\usepackage{diagbox}

\usepackage{algorithm}
\usepackage{algorithmic}
\usepackage{amssymb}
\usepackage{amsmath}
\usepackage{arydshln}
\usepackage{multirow}
\usepackage{color}
\newcommand{\tabincell}[2]{\begin{tabular}{@{}#1@{}}#2\end{tabular}}

\definecolor{cvprblue}{rgb}{0.21,0.49,0.74}
\usepackage[pagebackref,breaklinks,colorlinks,allcolors=cvprblue]{hyperref}

\def\paperID{4872} 
\def\confName{CVPR}
\def\confYear{2025}

\title{Remote Photoplethysmography in Real-World and Extreme Lighting Scenarios}

\author{Hang Shao$^{1}$, Lei Luo$^{1}$\textsuperscript{*}, Jianjun Qian$^{1}$, Mengkai Yan$^{1}$, Shuo Chen$^{2}$, Jian Yang$^{1}$\thanks{\scriptsize{Corresponding authors.}}\\
$^{1}$PCA Lab\thanks{\scriptsize{Key Lab of Intelligent Perception and Systems for High-Dimensional Information of Ministry of Education, School of Computer Science and Engineering.}}, Nanjing University of Science and Technology, $^{2}$Nanjing University, China\\
{\tt\small \{shaohang,cslluo,csjqian,ymk,csjyang\}@njust.edu.cn, shuo.chen.ya@foxmail.com}
}

\begin{document}
\maketitle

\begin{abstract}
	Physiological activities can be manifested by the sensitive changes in facial imaging. While they are barely observable to our eyes, computer vision manners can, and the derived remote photoplethysmography (rPPG) has shown considerable promise. However, existing studies mainly rely on spatial skin recognition and temporal rhythmic interactions, so they focus on identifying explicit features under ideal light conditions, but perform poorly in-the-wild with intricate obstacles and extreme illumination exposure. In this paper, we propose an end-to-end video transformer model for rPPG. It strives to eliminate complex and unknown external time-varying interferences, whether they are sufficient to occupy subtle biosignal amplitudes or exist as periodic perturbations that hinder network training. In the specific implementation, we utilize global \text{interference sharing, subject} background reference, and self-supervised disentanglement to eliminate interference, and further guide learning based on spatiotemporal filtering, reconstruction guidance, and frequency domain and biological prior constraints to achieve effective rPPG. To the best of our knowledge, this is the first robust rPPG model for real outdoor scenarios based on natural face videos, and is lightweight to deploy. Extensive experiments show the competitiveness and performance of our model in rPPG prediction across datasets and scenes.
\end{abstract}

\section{Introduction}

Heart rate (HR), heart rate variability, respiratory rate, blood pressure, and blood oxygen saturation are significant vital parameters, and their remote sensing can provide important evidence for many tasks for clinical purposes and intelligent applications. Remote photoplethysmography (rPPG) aims to capture blood volume pulse (BVP) signals containing the above bio-indicators based on facial videos \cite{wang_tbme17, shao_tim24}. With the popularization of commercial cameras and mobile communication devices, rPPG technique will play an irreplaceable role in health self-examination, anti-spoofing, driver fatigue diagnosis, and DeepFake in the foreseeable future.

\begin{figure}[t]
	\centering
	\includegraphics[width=1\columnwidth]{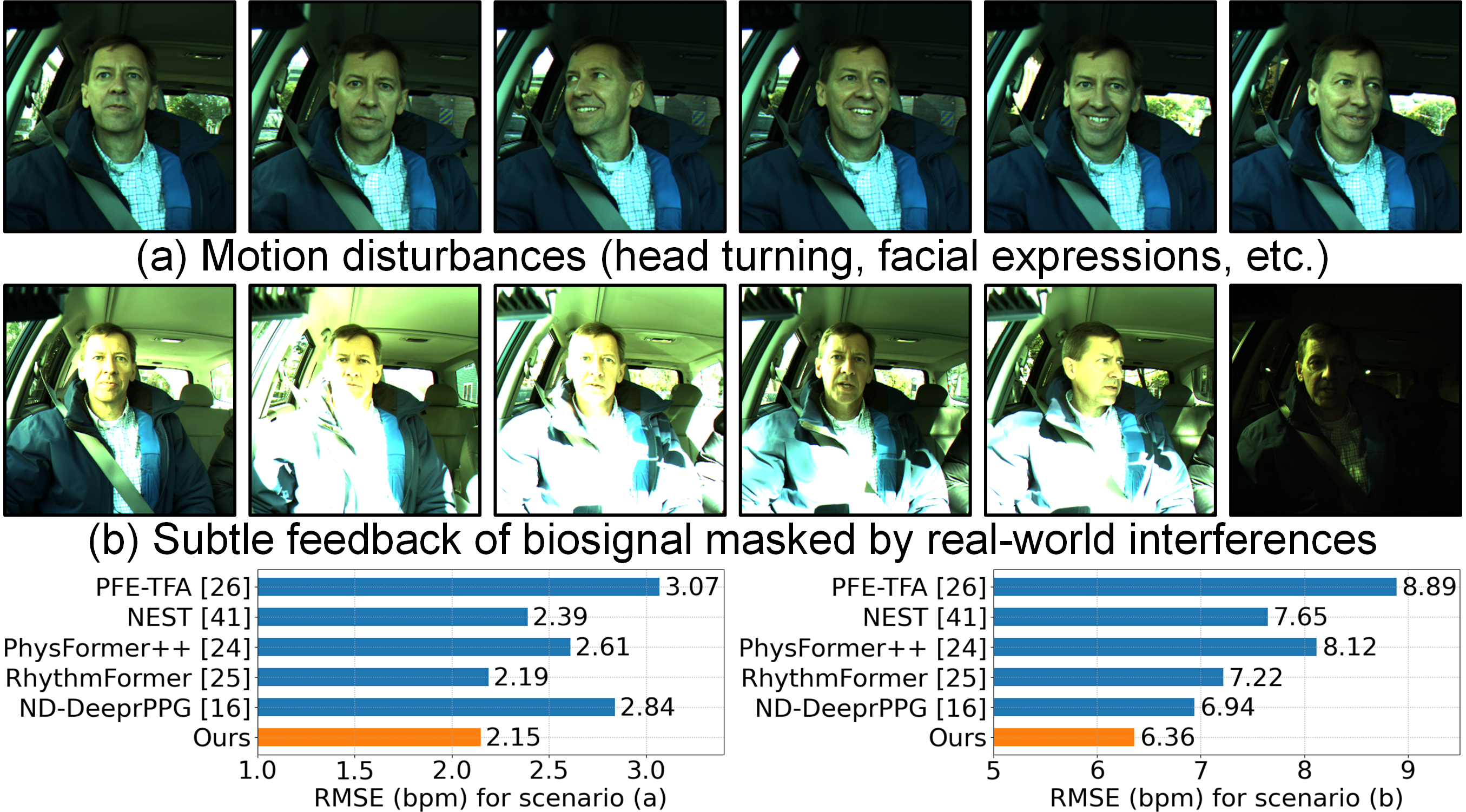}
	\caption{Existing learning-based rPPGs have shown promise under static illuminations (a), but they are not effective in real-world scenarios (b). Performances of different methods can be compared intuitively (under the MR-NIRP-DRV \cite{Nowara_tits22} dataset).}
	\label{img01}
\end{figure}

The contraction and relaxation of cardiac alter the instantaneous flow and velocity of red blood cells in subcutaneous capillaries, thereby changing facial skin imaging. Although these contents are difficult to be directly perceived by naked eyes, camera can capture the absorption and scattering of incident light caused by variations in superficial living tissues \cite{haan_tbme13, shao_tnnls24}. On these bases, rPPG summarizes biosignal clues by comparing skin color between frames. Early related studies implemented rPPG based on manual features \cite{Tulyakov_cvpr16, Pilz_cvpr18}. Subsequently, deep learning-based methods showed stronger robustness in motion and heterogeneous videos \cite{yu_cvpr22, sun_pami24}.

However, similar to the bottleneck of the traditional blind source separation methods which can demodulate as long as biosignal meets certain conditions, existing learning-based rPPG methods are also vulnerable in many cases, especially in the following aspects: 1) \textbf{Insufficient ability to remove extreme light interference.} Facial color changes caused by heart activity are relatively asthenic. When external light interference is sufficient to mask such changes, existing rPPG methods that mine clues from skin region-of-interest (ROI) can easily fail \cite{liu_nips20, yue_tim22}. 2) \textbf{Weak ability to resist periodic interference.} Cardiac signals have quasi-periodic characteristics, and many existing long-term BVP perception models are developed based on this. However, when the external interference also shows a certain time-varying and rhythmicity, it is more difficult to unravel \cite{lee_cvpr23, liu_arxiv24}. 3) \textbf{Still need to employ additional expensive equipment in practice.} For the unknown complex interference, some studies choose to introduce near-infrared (NIR) imaging as multimodal input, but this leads to the community return to the old routine of complicated devices and high costs \cite{chiu_tim23, gong_tim24}. 4) \textbf{Difficulty in model deployment.} Existing rPPG networks, especially those for the noise disentanglement, continue to accumulate heavy modules and complex pipelines, making them difficult to deploy on flexible hardware \cite{zhang_tcsvt24, liu_tip24}. In summary, in the real-world (such as driving scenarios) where the above challenges coexist, operating learning-based rPPG models rarely perform satisfactorily (as shown in Fig.~\ref{img01}).

In this paper, we propose a novel rPPG framework based on vision transformer, which can be applied to a variety of application scenes with drastic illumination and complex in-the-wild interference, and can be better adapted to terminals due to its lightweight layout. Specifically, to prevent rPPG waveforms from being overwhelmed by external interference, our model exploits the common property of facial skin and subject background disturbances, takes the background interference as a reference, discouples it in a self-supervised manner, and eliminates its influence in the foreground. In this process, we increase the frame window length to make full utilize of BVP quasi-periodic features that are different from extreme lighting obstructions. Meanwhile, we use the biosignal frequency domain and biological prior constraints to prevent the influence of interference with rhythmic characteristics that are hidden in too long a temporal period, and use spatiotemporal filtering and reconstruction to eliminate sensitive noise and better guide network training. It is worth noting that all our learning support is based on facial optical reflectance model and feature mapping correlation calculation, without introducing additional imaging modalities and too many branch modules, so the deployment cost is lower. We conduct experiments on multiple mainstream datasets, including BUAA-MIHR (illuminance state variation \cite{xi_fg20}), VIPL-HR (indoor real state \cite{niu_tip20}), MR-NIRP-IND (motion \cite{Nowara_cvpr18}), and MR-NIRP-DRV (outdoor driving moment \cite{Nowara_tits22}). Sufficient verification results demonstrate that our approach shows significant advantages in actual open and challenging scenarios. In summary, the main contributions include:

\begin{itemize}
	\item{We pioneer rPPG learning from facial videos in outdoor variable illumination, and propose a targeted paradigm. It relies merely on RGB imaging instead of various infrared spectra, aiming to extend rPPG to more realistic domains.}
	
	\item{Our model is guided by global interference sharing, background perturbation reference, self-supervised disentanglement, and long-term frequency and physiological prior constraint strategies, which can provide a robust solution for resisting unknown complex obstructions.}
	
	\item{Our model is lighter than existing noise disentanglement methods, can be deployed more flexibly, and is more suitable for in-the-wild platforms and real-world tasks.}
	
	\item{We compare with the latest representative rPPG methods and demonstrate that our model achieves the state-of-the-art results in both extreme lighting and static scenes.}	
\end{itemize}

\section{Related works}

\subsection{Learning-based rPPG in daily life}

Recently, researchers have developed many learning-based visual rPPG networks, which have continuously improved task-oriented performances in facial ROI attention \cite{liu_nips20, liu_wacv23}, spatiotemporal enhancement \cite{gupta_wacv23, zou_mamba24}, and long-term time perception \cite{yu_ijcv23, zou_24}. However, capabilities of these methods in open conditions are worrying. Currently, rPPG for actual scenes mainly solves the problems of subject motion robustness \cite{li_aaai23, Cantrill_cvpr24} and shallow denoising \cite{lu_cvpr21, du_cvpr23, savic_fg24}, and rarely involves extreme real-time changes in illumination. Even the most widely used VIPL-HR dataset \cite{niu_tip20}, which is generally considered to be the closest to real states, is not exception. Although some works have begun to focus on outdoor lighting, they all require the exploit of auxiliary infrared devices and have limited surface feature abilities \cite{Kurihara_tip21, xu_tim23}. Infrared imaging with a specific wavelength has advantages in counteracting visible light variations with a wide wavelength distribution, but according to the skin optical reflection model, facial imaging caused by biological activities is mainly due to the reflection of light by red blood cells and the absorption of melanin. At the cellular level, short-wave reflection is more effective. Therefore, its effectiveness in rPPG imaging demands to be in-depth evaluated \cite{chen_eccv18}. Meanwhile, narrowband waves contain less content, and the involvement of their equipment will make the deployment cost of detection more expensive. Therefore, rPPG in this field demands to be solved urgently to further serve downstream tasks \cite{zhang_spl23, Nguyen_24}.

\subsection{Self-supervised learning for rPPG}

Self-supervised learning is crucial for eliminating unknown interferences that affect rPPG. Initially, related approaches were mainly used to segment unlabeled skin \cite{bobbia_prl19} and solve annotation scarcity and misalignment \cite{shao_ijcai24}. \text{Gideon et al.} \cite{gideon_iccv21} used the quasi-periodicity of BVP to improve the network's ability to mine biosignal by learning its power spectral density (PSD). \text{Shao et al.} \cite{shao_tcsvt24} used the masked autoencoding paradigm to avoid the distortion of rPPG fragile amplitude by redundant spatial information. \text{Liu et al. \cite{liu_tmm24}} improved it to the multi-mapping parallelism. For unseen distributions, \text{Sun et al.} \cite{sun_pami24} explored conflict domains between different subjects through harmonic phase and hyperplane optimization, thereby improving \text{the model generalization.} \text{Lu et al.} \cite{lu_cvpr23} aligned the spatiotemporal mapping of source and target domains based on maximizing the coverage and diversity of neural structures to avoid the abnormal activati-on caused by sample inhomogeneity \cite{wang_cvpr22}. \text{Liu et al.} \cite{liu_tip24, liu_fg20} used adversarial canonical correlation analysis to separate background noise from original \text{cardiac features. Moreover,} researchers have proposed approaches such as frequency reconstruction \cite{yue_pami23, huang_24}, conductive inference \cite{lee_eccv20, birla_wacv23, Benezeth_24, savic_arxiv24}, multi-modality fusion \cite{park_tim22}, and pseudo-label recombination \cite{mcduff_nips22, li_iccv23}. However, they often ignore the impact of environmental changes on facial color imaging, especially when these disturbances also show certain rhythmic characterizations.

\section{Methodology}

There are many challenges to overcome in order to make a learning-based rPPG framework work effectively outdoors. Therefore, when designing the end-to-end network, we had to account for interference from motion and general noise, as well as extremely complex lighting variations.

\subsection{Basic general interference elimination}

\begin{figure}[t]
	\centering
	\includegraphics[width=1\columnwidth]{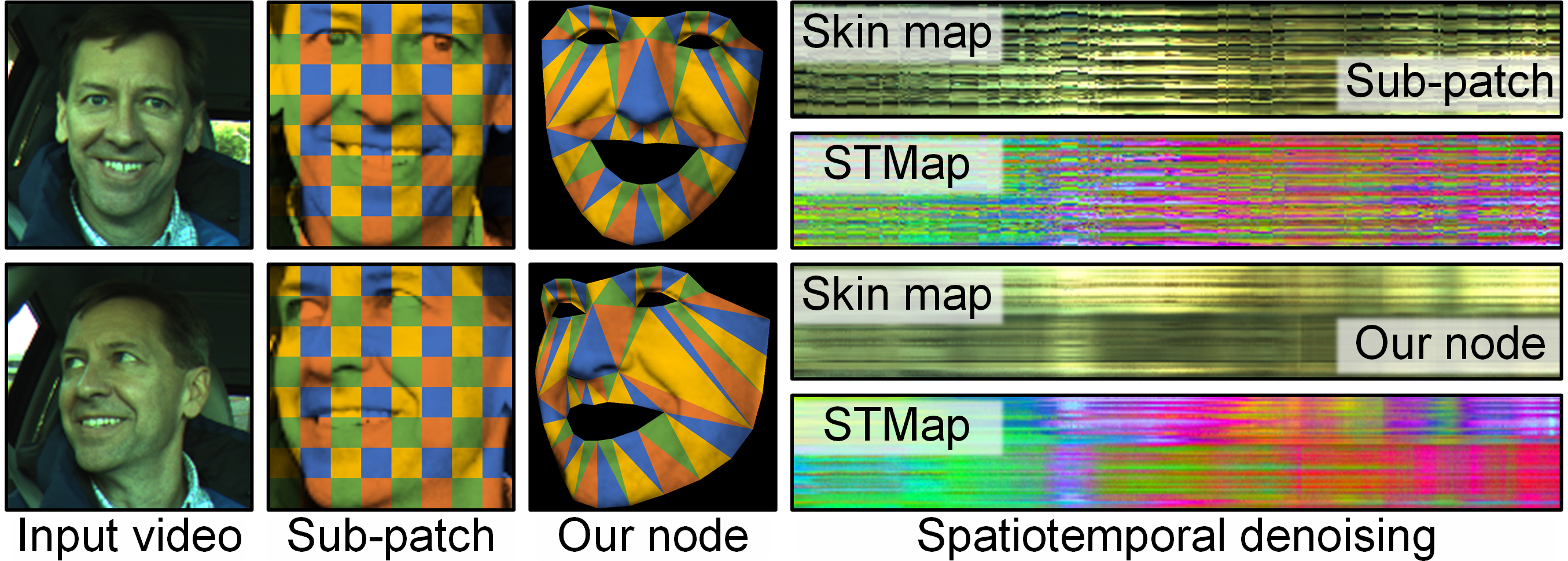}
	\caption{Our facial skin feature extraction method and the STMap built on it can enhance subtle color changes and spikes more finely than traditional sub-patch sliding window denoising algorithms.}
	\label{img02}
\end{figure}

We follow the boosting design. First, to remove the general noise and for the sake of subsequent lightweight layout and efficient training, we transform the input RGB video into a YUV spatiotemporal map (STMap) \cite{niu_tip20} in the color space. This operation converts single frame into a one-dimensional array within the color channel, and constructs a new image based on the temporal length of video clips. Although there have been many works exploring the benefits of such multi-scale pooling style ideas for the biosignal denoising \cite{niu_fg19, ling_prcv22} and their contributions to embedding the latest frameworks into rPPG \cite{lu_cvpr23, savic_arxiv24}, their latent is easily distorted when dealing with movements. Therefore, we improve that to a node graph manner based on facial landmarks. We split and divide the face skin into multiple non-overlapping regions according to its landmarks, which are consistent with the distribution of subcutaneous tissue units rather than shallow rectangle sub-patches. We calculate the average pixel value in each independent region. For an input natural video sequence $\mathbf{V}_{\rm in} \! \in \! \mathbb{R}^{3 \times T \times H \times W}$ with $T$ frames, its facial STMap can be expressed as: $\mathbf{V}_{\rm face} \! \in \! \mathbb{R}^{3 \times L \times T}$, where $H$ and $W$ are the height and width of frames, and $L$ is the number of landmark points. As shown in Fig.~\ref{img02}, this can make our STMap immune to motion and behavior interference.

HR transitions smoothly in the most cases. Therefore, in a short time period, rPPG fragments manifest similarity and quasi-periodicity, specifically in BVP frequency, amplitude, systolic point, diastolic point, and dicrotic notch. Based on these features, increasing the temporal dimension of STMap is conducive to strengthen long-term contextual interaction of biosignals, thereby improving the trained model robustness. The very recently produced vision transformer framework is also developed for related linear complexity. Therefore, we chose the Swin video transformer \cite{cao_eccv22} with strong long-term capabilities as our backbone architecture. 

However, blindly increasing the temporal dimension will cause the preprocessing of STMap to fall into many uncertain factors. Especially when facing severe disturbance with an amplitude much greater than skin color changes, existing normalization methods of STMap will further magnify the interference \cite{shao_ijcai24}. To this end, we propose a novel augmentation algorithm. The camera frame rate available for rPPG is usually 20 to 30 frames per second (fps), and HR is generally in the range of 40 to 240 beats per minute (bpm) \cite{Jeanningros_24}. In view of this important physiological knowledge, we perform a biological prior-based sliding window enhancement (BioSE) on the raw STMap. Specifically, 1) To increase the independent representation of each organ unit, we calculate the color variations over time in each node region based on face landmarks. 2) To avoid the global normalization weakening in BVP imaging, we calculate sliding within the preset window length based on the biological prior to amplify subtle color differences and spike features. 3) To make the edges of windowed signal continuous and smooth, we adopt a parallel manner with multiple different starting positions. The imaging ${v}^{\rm face}$ in each independent unit is expressed:

\begin{equation}
	{\rm BioSE} ~ {v}_{t, n-t+{s}_{\rm norm}}^{\rm face} = ({v}_{n}^{\rm face} - {\rm min}_{t}) / ({\rm max}_{t} - {\rm min}_{t}) ,
\end{equation}

\noindent where $t \! - \! {s}_{\rm norm} \! < \! n \leq t$, $t$ is the current discrete time, $s_{\rm norm}$ is the span of the sliding window, $n$ is the discrete time numbers in the sliding window, ${v}_{i}^{\rm face}$ is the $i$-th processed ${v}^{\rm face}$, ${\rm BioSE} ~ {v}_{t, n-t+s_{\rm norm}}^{\rm face}$ is the skin imaging variation to which the $(n \! - \! t \! + \! {s}_{\rm norm})$-th proposed method (BioSE) is applied at the $t$-th, ${\rm max}_{t}$ and ${\rm min}_{t}$ are the max and min deviations on the $t$-th sliding window. According to the facial skin optical feedback \cite{liu_nips20}, to prevent implicit relations between the epidermal specular reflection and the subcutaneous tissue diffuse one from being destroyed in the above processes, we connect the map through BioSE with the raw facial STMap. That is also because R, G, and B act differently on the face. While the G channel performs well under the indoor LED lamps, its effect on real outdoor sunlight remains to be studied. Then, a new STMap ($\mathbf{S}_{\rm face} \! \in \! \mathbb{R}^{6 \times L \times T}$) with raw details and light-insensitive biosignals can be obtained:

\begin{equation}
	\mathbf{S}_{\rm face} = {\rm Concate} \Big( \mathbf{V}_{\rm face}, \frac{255}{N} \sum_{i=1}^{N} {\rm BioSE} \big(\mathbf{V}_{{\rm face}_{i}} \big) \Big) ,
\end{equation}

\noindent where $N$ is the number of BioSE performed with different starting frames, and ${\rm Concate}$ is the concatenate in NumPy.

\subsection{Boosted complex interference disentanglement}

To further improve the advantages of our model under complex conditions, which is also the core distinguish from current rPPGs, we make full use of the theory of global intra-frame interference sharing. For example, in a driving scene, when a vehicle travels through a city, the entering light will change at any time due to car turns, building occlusions, etc. However, the interference of the main object of the person, the kernel skin ROI of the face, and the background is consistent. Therefore, on the basis of remapping the enhanced STMap based on the facial skin node set, we also construct a background STMap ($\mathbf{S}_{\rm back}$), which is the \text{region outside the} bounding box formed by the facial landmarks. To ensure the subsequent linear embedding and patch merging, we project the face background STMap in the same way as $\mathbf{V}_{\rm face}$, except that each frame is average pooled in $L$ rectangular windows so that $\mathbf{S}_{\rm back}$ and $\mathbf{S}_{\rm face}$ have the same tensor shape.

Subsequently, we use the background perturbation as the reference and mine the potential temporal similarity in $\mathbf{S}_{\rm face}$ and $\mathbf{S}_{\rm back}$ to disentangle strong interference. Based on this, we hope to embed them into a unified framework to capture their commonalities while not ignoring their individuality. Therefore, we need to enable our model pre-form a concept of the calculated features in advance, but not to make them too significant to avoid overfitting. Based on the above discussions, we additionally construct a global STMap ($\mathbf{S}_{\rm glob}$, shaped like $\mathbf{S}_{\rm face}$), which is obtained by dividing the overall input video frame into $L$ blocks and computing their means. Using the global map as the initial input of learning pipeline can be regarded as training in a way that actively reduces the signal-to-noise ratio of the modeling subject, avoiding premature fitting of fine-grained concepts. Meanwhile, we improve the traditional one-dimensional BVP regression to the spatiotemporal reconstruction of the STMap by the stacked BVP signals to increase the temporal interaction of different STMap sub-blocks, and design the model as a U-shaped structure. As far as we know, it is also the front explorations of U-shaped transformer in biosignal measurement. In addition, the expansion of mapping relations enables our model to be trained directly from scratch on rPPG datasets.

While reconstructing the global frame map $\mathbf{S}_{\rm glob}$ at the spatiotemporal level using the U-shaped video transformer, we extract the corresponding implicit feature from the middle layer of the architecture. Since this feature comes from the overall video with unknown noise, we let it as a coarse-grained feature $\mathbf{F}_{\rm coar}$. Correspondingly, we feed $\mathbf{S}_{\rm face}$ and $\mathbf{S}_{\rm back}$ together into the encoder of the U-shaped backbone to obtain foreground feature $\mathbf{F}_{\rm fore}$ and background one $\mathbf{F}_{\rm back}$. They satisfy: $\mathbb{R}^{C \times L' \times T'}$, where $C$, $L'$, and $T'$ are the new dimensions after channel excitation and STMap downsampling. Since reflectance and material of diverse body background regions are not the same, we do not know the distribution of background variations and cannot pre-define or assume them. Therefore, we employ a self-supervised manner to query the foreground feature $\mathbf{F}_{\rm fore}$ for its response using the background tensor $\mathbf{F}_{\rm back}$ as a reference to capture sub-blocks with similar temporal representations in foreground and background maps. To make our architecture compatible with general rPPG tasks, after obtaining the similarity score mapping, we do not adopt the strategy of setting hyperparameters and selecting the first few high-scoring blocks as interference removal. In response, global adaptive computation is utilized instead. Specifically, we construct a newly fine-grained tensor $\mathbf{F}_{\rm fine}$, which considers the influence and weights of all sub-regions within the face:

\begin{figure}[t]
	\centering
	\includegraphics[width=1\columnwidth]{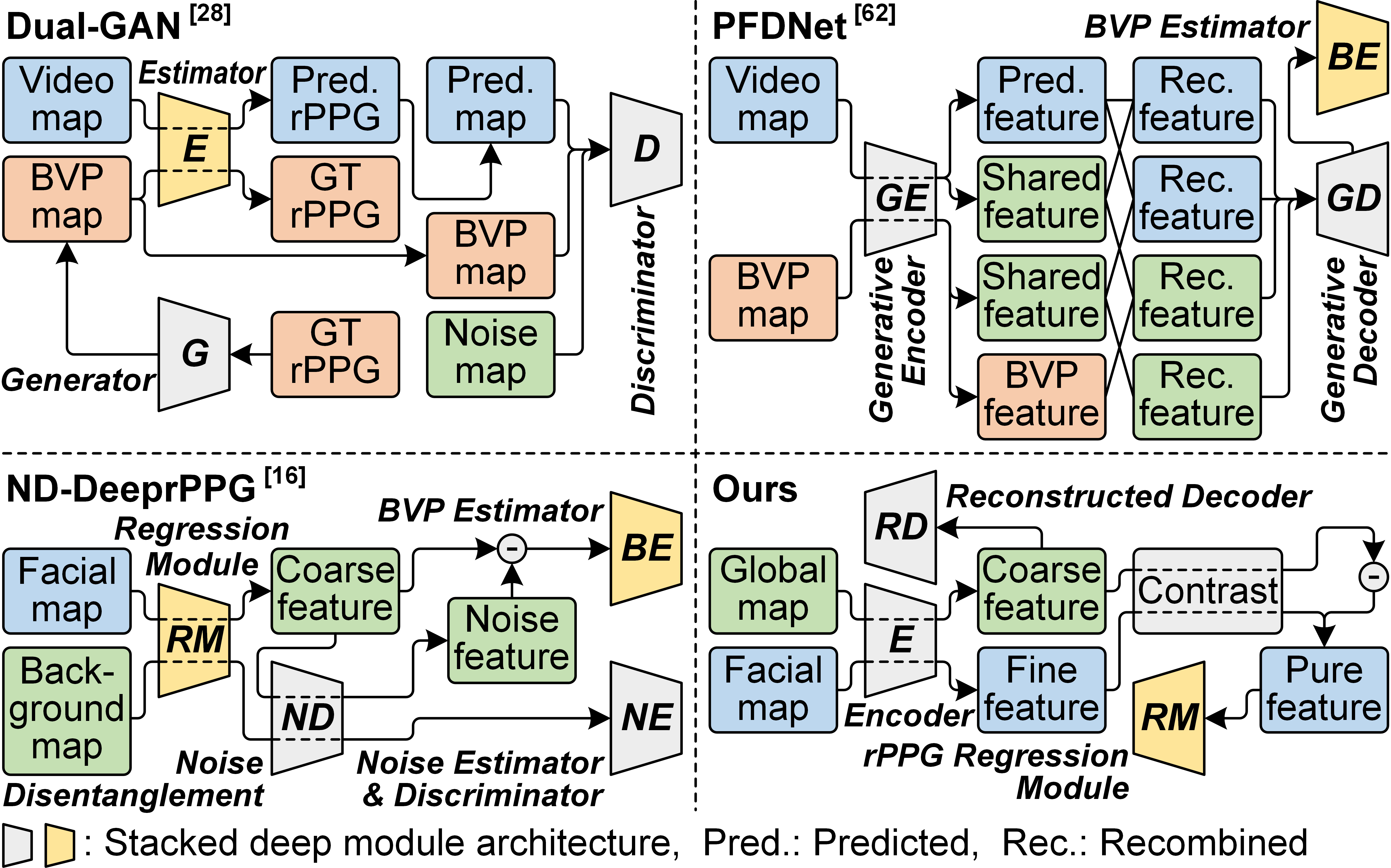}
	\caption{Comparison of our method with existing rPPG interference disentanglement models that are representative in paradigm.}
	\label{img03}
\end{figure}

\begin{equation}
	\mathbf{F}_{\rm fine} = \Big( 1 - {\rm Softmax} \big( \mathbf{F}_{\rm back} \cdot ( \mathbf{F}_{\rm fore} )^\mathsf{T} ~ \big) \Big) \cdot \mathbf{F}_{\rm fore} .
\end{equation}

After that, we guide interference disentanglement by latent contrastive learning. We construct a contrast constraint function $\mathcal{L}_{\rm c}$ to calculate the temporal relation between different facial sub-regions and background sub-patches. According to the irrelevance coefficient, the facial sub-regions that are irrelevant to the background pattern are pulled away, and the relevant ones are pushed away:

\begin{figure*}[t]
	\centering
	\includegraphics[width=2\columnwidth]{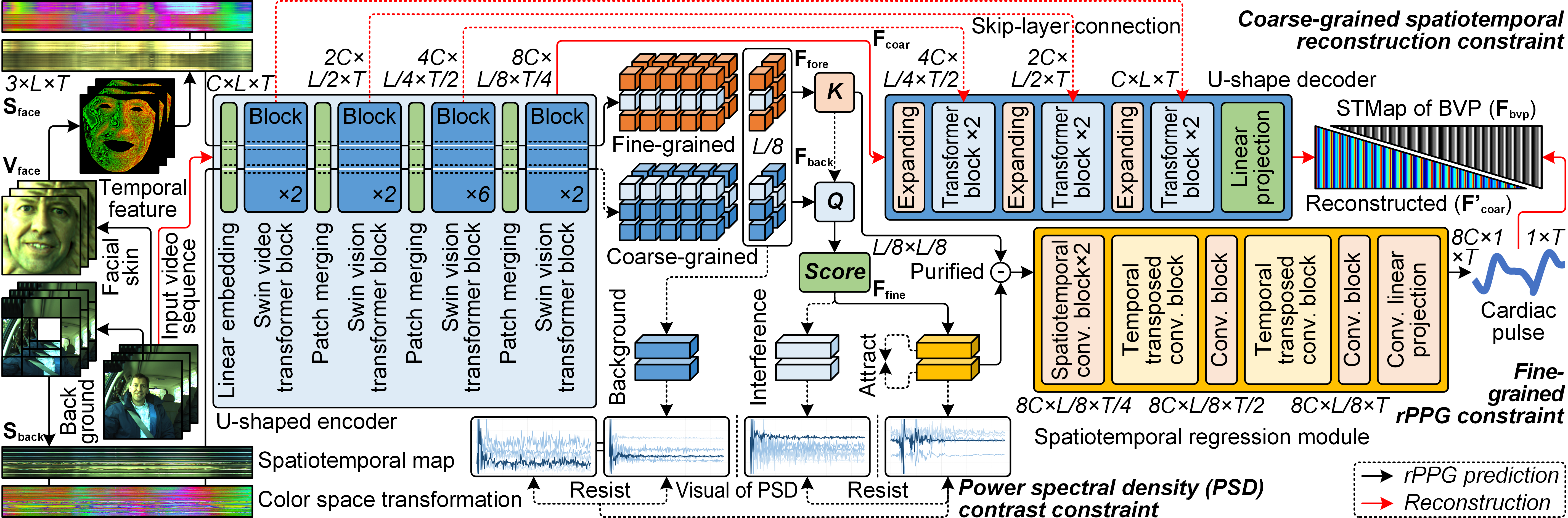}
	\caption{The framework of our end-to-end time-varying interference disentanglement network, which consists of a U-shaped transformer module for coarse-grained spatiotemporal reconstruction and an rPPG prediction module for fine-grained BVP waveform regression.}
	\label{img04}
\end{figure*}

\begin{equation}
	\mathcal{L}_{\rm c} = {\rm log} \bigg( \frac{\sum_{i_1=1}^{L'} \sum_{\substack{i_2=1 \\ i_1 \neq i_2}}^{L'} {\rm exp} \big( {D}(\mathbf{F}'_{{\rm fine}_{i_1}}, \mathbf{F}'_{{\rm fine}_{i_2}})/\tau \big) }{\sum_{i=1}^{L'} \sum_{j=1}^{L'} {\rm exp} \big( D(\mathbf{F}'_{{\rm fine}_{i}}, \mathbf{F}'_{{\rm back}_{j}})/\tau \big) } + 1 \bigg) ,
\end{equation}

\noindent where $D$ is the mean square error of two tensors with respect to time, $\mathbf{F}'_{{\rm fine}_{i}}$ and $\mathbf{F}'_{{\rm back}_{i}}$ are the tensors of $\mathbf{F}_{\rm fine}$ and $\mathbf{F}_{\rm back}$ at the $i$-th sub-block, and $\tau$ is the temperature hyper-parameter which is set to 0.08 following \cite{yue_pami23}. Besides, the global reconstruction features do not participate in the disentanglement calculation. Their purpose is to continuously iterate network parameters, increase the model's abilities to express fine-grained and background features, and prevent the encoder from overfitting fine-grained features or fitting noise. At this point, still taking the city driving scenario as an example, the regularity of street lights and roadside trees will make the external interference periodic when the vehicle passes by. Studies have proven that rPPG biosignals obtained from different facial sub-regions often show similar power spectrum \cite{sun_eccv22}. Therefore, to avoid above influences, we adopt power spectral density (PSD) of imaging features as the target of the distance metric for $D$ calculation.

It is worth mentioning that the current methods that consider illumination changes usually require the assistance of infrared rays, but according to the surface optical reflection theory, the feedback of cardiac activity is the variation that occurs when natural light passes through the pulsating cell flow in the capillaries, not thermal radiation, and there is an essential difference between the two. Therefore, decoupling based on visible light is the key to solving this problem. In addition, it can be seen that our computing is mainly based on the facial reflectance and feature correlation calculation, unlike existing disentanglement methods that rely on generative networks \cite{lu_cvpr21, tran_cvpr17, liu_cvpr18, song_jbhi21} or adversarial learning \cite{liu_tip24, niu_eccv20}, which requires the introduction of too many heavy modules. Therefore, it is more lightweight. Furthermore, we present differences between ours and existing modules (Dual-GAN \cite{lu_cvpr21}, PFDNet \cite{Liu_jbhi23}, ND-DeeprPPG \cite{liu_tip24}) in Fig.~\ref{img03}. Specifically, although PFDNet is based on rPPG for atrial fibrillation screening, its core is still the purification of BVP waveforms. As for ND-DeeprPPG, its core is to equip a shallow model with a noise disentanglement module. The similarity between our algorithm and it is that we both take advantage of the consistency of skin (foreground)-environment (background) noise. The main difference is that it needs to build an external discriminator to judge the HR counted by the environment as false and the value counted by the skin as true. We directly adopt BVP with the same \text{time dimension as} the video for spectrum consistency comparison, which is more in line with the requirements of long sequence regression.

\subsection{Overall rPPG prediction and optimization}

For the global $\mathbf{F}_{\rm coar}$ which dose not participate in disentanglement, it is then reconstructed by the U-shaped decoder, and the spatiotemporal reconstruction loss $\mathcal{L}_{\rm r}$ is counted by stacking the ground truth (GT) biological signal $B_{\rm g}$, so that our model can increase the perception of BVP from coarse to fine: $\mathcal{L}_{\rm r} = || \mathbf{F}'_{\rm coar} - \mathbf{F}_{\rm bvp}||_{2}$, where $\mathbf{F}'_{\rm coar}$ is the feature of $\mathbf{F}_{\rm coar}$ expanded by the decoder projection, and $\mathbf{F}_{\rm bvp}$ is the single channel STMap stacked by $B_{\rm g}$. For $\mathbf{F}_{\rm fine}$ after eliminating the interference, we construct a separate regression module based on temporal deconvolution in addition to the spatiotemporal reconstruction to predict the remote biosignal $B_{\rm p}$. Specifically, it consists of six groups of blocks including convolutional layers, activation, and batch normalization. Without varying the node scale of $L'$, the downsampled time dimension is transposed to the input level layer by layer. The loss function $\mathcal{L}_{\rm p}$ that focuses on temporal series regression and spike protection at this stage is:

\begin{equation}
	\mathcal{L}_{\rm p} \! = \! 1 \! - \frac{T \sum\nolimits_{i=1}^{T} {B_{\rm p}}_{i} {B_{\rm g}}_{i} - \sum\nolimits_{i=1}^{T} {B_{\rm p}}_{i} \sum\nolimits_{i=1}^{T} {B_{\rm g}}_{i}} { \! \sqrt{ \! \Big( \! T \! \sum\limits_{i=1}^{T} \! {B_{\rm p}}_{i}^{2} \! - \! ( \sum\limits_{i=1}^{T} \! {B_{\rm p}}_{i} ) ^{2} \! \Big) \! \Big( \! T \! \sum\limits_{i=1}^{T} \! {B_{\rm g}}_{i}^{2} \! - \! ( \sum\limits_{i=1}^{T} \! {B_{\rm g}}_{i} )^{2} \! \Big) \! }} .
\end{equation}

\begin{table*}[t]
	\begin{center}
		\caption{Comparative results (bpm) of remote HR estimation on MR-NIRP-IND, MR-NIRP-DRV, VIPL-HR, and BUAA-MIHR datasets, where $\downarrow$ means the smaller the better, $\uparrow$ is the bigger the better, the best result is \textbf{bolded}, and the second best is \underline{underlined} (the same below).}
		\resizebox{\linewidth}{!}{
			\begin{tabular}{c|l|c|ccc|ccc|ccc|ccc}
				\hline
				\multicolumn{2}{l|}{\multirow{2}{*}{rPPG Methods / Venues}} & \multirow{2}{*}{\diagbox{Train}{Test}} & \multicolumn{3}{c|}{MR-NIRP-IND \cite{Nowara_cvpr18}} & \multicolumn{3}{c|}{MR-NIRP-DRV \cite{Nowara_tits22}} & \multicolumn{3}{c|}{VIPL-HR \cite{niu_tip20}} & \multicolumn{3}{c}{BUAA-MIHR \cite{xi_fg20}} \\ \cline{4-15}
				\multicolumn{1}{l}{\multirow{2}{*}{}} & {} & {} & MAE$\downarrow$ & RMSE$\downarrow$ & ${\rho}$$\uparrow$ & MAE$\downarrow$ & RMSE$\downarrow$ & ${\rho}$$\uparrow$ & MAE$\downarrow$ & RMSE$\downarrow$ & ${\rho}$$\uparrow$ & MAE$\downarrow$ & RMSE$\downarrow$ & ${\rho}$$\uparrow$ \\ \hline
				
				\multicolumn{2}{l|}{{CHROM}\cite{haan_tbme13} / {\em IEEE TBME 2013}} & \multicolumn{1}{c|}{n/a} & {~~6.84} & {~~8.41} & {0.32} & {14.52} & {17.41} & {0.18} & {11.07} & {17.99} & {0.27} & ~~{6.09} & ~~{8.29} & {0.51} \\ \hdashline
				
				\multicolumn{2}{l|}{{POS}\cite{wang_tbme17} / {\em IEEE TBME 2017}} & \multicolumn{1}{c|}{n/a} & {~~5.52} & {~~6.85} & {0.40} & {12.75} & {15.36} & {0.34} & {11.50} & {17.20} & {0.30} & ~~{5.04} & ~~{7.12} & {0.63} \\ \hdashline
				
				\multicolumn{2}{l|}{{LGI}\cite{Pilz_cvpr18} / {\em CVPR 2018}} & \multicolumn{1}{c|}{n/a} & {~~8.95} & {11.01} & {0.39} & {12.90} & {15.51} & {0.31} & {12.84} & {19.02} & {0.29} & ~~{6.97} & {11.33} & {0.42} \\ \hline
				
				\multirow{11}{*}{\rotatebox{90}{CNN-Based rPPG Methods}} & \tabincell{l}{DeepPhys \cite{chen_eccv18} \\ \textit{ECCV 2018}} & \multicolumn{1}{c|}{\tabincell{c}{MR-NIRP \\ VIPL-HR}} & \tabincell{c}{~~3.11 \\ ~~6.58} & \tabincell{c}{~~4.44 \\ ~~9.16} & \tabincell{c}{0.74 \\ 0.52} & \tabincell{c}{13.22 \\ 10.51} & \tabincell{c}{18.39 \\ 12.71} & \tabincell{c}{0.43 \\ 0.56} & \tabincell{c}{15.53 \\ {11.00}} & \tabincell{c}{17.48 \\ {13.80}} & \tabincell{c}{0.41 \\ {0.11}} & ~~{4.78} & ~~{6.74} & {0.69} \\ \cdashline{2-15}
				
				{} & \tabincell{l}{TS-CAN \cite{liu_nips20} \\ \textit{NeurIPS 2020}} & \multicolumn{1}{c|}{\tabincell{c}{MR-NIRP \\ VIPL-HR}} & \tabincell{c}{~~2.49 \\ ~~6.57} & \tabincell{c}{~~3.89 \\ ~~9.25} & \tabincell{c}{0.75 \\ 0.57} & \tabincell{c}{12.70 \\ 10.37} & \tabincell{c}{18.03 \\ 12.65} & \tabincell{c}{0.47 \\ 0.61} & \tabincell{c}{12.04 \\ ~~{9.39}} & \tabincell{c}{15.12 \\ {14.59}} & \tabincell{c}{0.45 \\ {0.21}} & ~~{4.84} & ~~{6.89} & {0.68} \\ \cdashline{2-15}
				
				{} & \tabincell{l}{Dual-GAN \cite{lu_cvpr21} \\ \textit{CVPR 2021}} & \multicolumn{1}{c|}{\tabincell{c}{MR-NIRP \\ VIPL-HR}} & \tabincell{c}{~~2.47 \\ ~~4.04} & \tabincell{c}{~~3.62 \\ ~~6.07} & \tabincell{c}{0.84 \\ 0.73} & \tabincell{c}{~~8.00 \\ ~~9.23} & \tabincell{c}{12.18 \\ 11.34} & \tabincell{c}{0.71 \\ 0.62} & \tabincell{c}{~~9.48 \\ ~~{4.93}} & \tabincell{c}{11.66 \\ ~~{7.68}} & \tabincell{c}{0.49 \\ {0.81}} & ~~{3.41} & ~~{5.23} & {0.84} \\ \cdashline{2-15}
				
				{} & \tabincell{l}{PFE-TFA \cite{li_aaai23} \\ \textit{AAAI 2023}} & \multicolumn{1}{c|}{\tabincell{c}{MR-NIRP \\ VIPL-HR}} & \tabincell{c}{~~2.81 \\ ~~1.87} & \tabincell{c}{~~4.57 \\ ~~3.67} & \tabincell{c}{0.84 \\ 0.77} & \tabincell{c}{~~5.34 \\ ~~7.82} & \tabincell{c}{~~8.92 \\ ~~9.99} & \tabincell{c}{0.73 \\ 0.64} & \tabincell{c}{~~8.20 \\ ~~{6.91}} & \tabincell{c}{11.22 \\ ~~{8.65}} & \tabincell{c}{0.64 \\ {0.65}} & ~~{1.29} & ~~{2.65} & {0.91} \\ \cdashline{2-15}
				
				{} & \tabincell{l}{NEST \cite{lu_cvpr23} \\ \textit{CVPR 2023}} & \multicolumn{1}{c|}{\tabincell{c}{MR-NIRP \\ VIPL-HR}} & \tabincell{c}{~~1.08 \\ ~~2.78} & \tabincell{c}{~~2.26 \\ ~~4.63} & \tabincell{c}{0.89 \\ 0.78} & \tabincell{c}{~~3.61 \\ ~~6.27} & \tabincell{c}{~~7.32 \\ ~~8.45} & \tabincell{c}{0.82 \\ 0.65} & \tabincell{c}{~~5.15 \\ ~~{5.52}} & \tabincell{c}{~~8.78 \\ ~~{7.96}} & \tabincell{c}{0.72 \\ {0.80}} & ~~{2.88} & ~~{4.69} & {0.89} \\ \cdashline{2-15}
				
				{} & \tabincell{l}{ND-DeeprPPG \cite{liu_tip24} \\ \textit{IEEE TIP 2024}} & \multicolumn{1}{c|}{\tabincell{c}{MR-NIRP \\ VIPL-HR}} & \tabincell{c}{~~0.66 \\ ~~1.74} & \tabincell{c}{~~1.45 \\ ~~\underline{3.28}} & \tabincell{c}{0.89 \\ 0.81} & \tabincell{c}{~~3.47 \\ ~~6.18} & \tabincell{c}{~~\underline{6.54} \\ ~~\underline{7.75}} & \tabincell{c}{\underline{0.85} \\ 0.68} & \tabincell{c}{~~5.08 \\ ~~{5.15}} & \tabincell{c}{~~\underline{7.92} \\ ~~{7.52}} & \tabincell{c}{\underline{0.75} \\ {0.78}} & ~~\underline{0.58} & ~~{1.81} & \textbf{0.95} \\ \hline
				
				\multirow{5}{*}{\rotatebox{90}{Transformers}} & \tabincell{l}{EfficientPhys \cite{liu_wacv23} \\ \textit{WACV 2023}} & \multicolumn{1}{c|}{\tabincell{c}{MR-NIRP \\ VIPL-HR}} & \tabincell{c}{~~1.37 \\ ~~1.80} & \tabincell{c}{~~4.81 \\ ~~6.19} & \tabincell{c}{0.81 \\ 0.76} & \tabincell{c}{~~3.67 \\ ~~6.07} & \tabincell{c}{12.28 \\ 11.53} & \tabincell{c}{0.81 \\ 0.64} & \tabincell{c}{~~5.25 \\ ~~{5.23}} & \tabincell{c}{11.81 \\ ~~{8.25}} & \tabincell{c}{0.71 \\ {0.72}} & ~~{1.43} & ~~{4.98} & {0.93} \\ \cdashline{2-15}
				
				{} & \tabincell{l}{PhysFormer++ \cite{yu_ijcv23} \\ \textit{IJCV 2023}} & \multicolumn{1}{c|}{\tabincell{c}{MR-NIRP \\ VIPL-HR}} & \tabincell{c}{~~0.64 \\ ~~1.79} & \tabincell{c}{~~1.39 \\ ~~3.77} & \tabincell{c}{0.89 \\ 0.81} & \tabincell{c}{~~3.56 \\ ~~5.78} & \tabincell{c}{~~7.59 \\ ~~8.65} & \tabincell{c}{0.83 \\ 0.66} & \tabincell{c}{~~5.09 \\ ~~{4.88}} & \tabincell{c}{~~8.97 \\ ~~{7.62}} & \tabincell{c}{0.71 \\ {0.80}} & ~~{0.93} & ~~{1.66} & \underline{0.94} \\ \cdashline{2-15}
				
				{} & \tabincell{l}{RhythmFormer \cite{zou_24} \\ {\em arXiv 2024}} & \multicolumn{1}{c|}{\tabincell{c}{MR-NIRP \\ VIPL-HR}} & \tabincell{c}{~~\underline{0.52} \\ ~~\underline{1.70}} & \tabincell{c}{~~\underline{1.10} \\ ~~3.43} & \tabincell{c}{\underline{0.91} \\ \underline{0.83}} & \tabincell{c}{~~\underline{3.44} \\ ~~\underline{5.40}} & \tabincell{c}{~~6.72 \\ ~~7.93} & \tabincell{c}{0.82 \\ \underline{0.72}} & \tabincell{c}{~~\underline{4.83} \\ ~~\underline{4.75}} & \tabincell{c}{~~8.05 \\ ~~\underline{7.49}} & \tabincell{c}{\underline{0.75} \\ \underline{0.82}} & ~~{0.67} & ~~\underline{1.57} & \underline{0.94} \\ \hline
				
				\multicolumn{2}{l|}{Ours} & \multicolumn{1}{c|}{\tabincell{c}{MR-NIRP \\ VIPL-HR}} & \tabincell{c}{~~\textbf{0.45} \\ ~~\textbf{1.43}} & \tabincell{c}{~~\textbf{0.91} \\ ~~\textbf{3.05}} & \tabincell{c}{\textbf{0.92} \\ \textbf{0.88}} & \tabincell{c}{~~\textbf{2.69} \\ ~~\textbf{4.97}} & \tabincell{c}{~~\textbf{6.03} \\ ~~\textbf{7.23}} & \tabincell{c}{\textbf{0.87} \\ \textbf{0.74}} & \tabincell{c}{~~\textbf{4.13} \\ ~~\textbf{4.52}} & \tabincell{c}{~~\textbf{7.61} \\ ~~\textbf{7.09}} & \tabincell{c}{\textbf{0.76} \\ \textbf{0.85}} & ~~\textbf{0.53} & ~~\textbf{1.20} & \textbf{0.95} \\ \hline			
		\end{tabular}}
		\label{tab1}
	\end{center}
\end{table*}

Furthermore, in terms of biological priors, based on the rule of 40 to 240 heartbeats per minute, we focus on the PSD frequency between 0.66 to 4 Hz in $\mathbf{F}_{\rm fine}$ in the calculation of $\mathcal{L}_{\rm c}$. In terms of further spatiotemporal context interaction, after $\mathbf{F}_{\rm coar}$ passes through U-shaped model, it provides hierarchical information by multi-scale skip-layer connection to guide reconstruction. At this time, the traditional window shifting of the Swin mechanism destroys skin temporal continuities. Therefore, we modify it to the spatial shifting. We use a random spatial transformation for the Swin unit composed of two transformer blocks instead of utilizing a mask, randomly reorganizing its $L$ dimension in the first block and vice versa in the another. Since capillaries in different facial sub-regions are relatively independent, this can increase the diversity of spatial topology and improve the generalization of our model. In summary, our overall loss $\mathcal{L}_{\rm total}$ is:

\begin{table}[t]
	\begin{center}
		\caption{Results for different lights and modalities involving NIR.}
		\resizebox{\linewidth}{!}{
			\begin{tabular}{l|c|ccc|ccc}
				\hline
				\multirow{2}{*}{rPPG Methods} & \multicolumn{4}{c|}{RGB Mode (RMSE$\downarrow$)} & \multicolumn{3}{c}{NIR Mode (RMSE$\downarrow$)} \\ \cline{2-8}
				\multirow{2}{*}{} & {Mo.} & {Day} & {Night} & {Ga.} & {Day} & {Night} & {Ga.} \\ \hline
				{TURNIP \cite{comas_icip21}} & {n/a} & {n/a} & {n/a} & {n/a} & {11.40} & {11.27} & {4.60} \\
				{AutoSparsePPG \cite{Nowara_tits22}} & {17.69} & {15.92} & {16.23} & \underline{2.90} & {11.80} & {11.20} & {5.10} \\
				{Wu et al. \cite{wu_tim22}} & {15.41} & {10.87} & {14.46} & {5.94} & ~~{9.88} & ~~\textbf{9.58} & \underline{3.93} \\
				{Chiu et al. \cite{chiu_tim23}} & ~~\underline{9.71} & ~~\underline{7.60} & \underline{10.11} & {4.46} & ~~\underline{8.82} & ~~\underline{9.95} & {4.09} \\ \hline
				{Ours} & ~~\textbf{7.30} & ~~\textbf{5.45} & ~~\textbf{7.51} & \textbf{2.15} & ~~\textbf{7.52} & {10.93} & \textbf{3.84} \\ \hline
		\end{tabular}}
		\label{tab2}
	\end{center}
\end{table}

\begin{equation}
	\mathcal{L}_{\rm total} = \alpha ~ \mathcal{L}_{\rm r} + \beta ~ \mathcal{L}_{\rm c} + \gamma ~ \mathcal{L}_{\rm p} ,
\end{equation}

\noindent where $\alpha$=0.5, $\beta$=0.5, and $\gamma$=1 in our setting studies. Moreover, we can make our model compatible with existing common pathways through rPPG by controlling the coefficients of $\beta$ and the Hadamard product in $\mathcal{L}_{\rm c}$ as a score map. 

Our total workflow is shown in Fig.~\ref{img04}. We mark the spatiotemporal reconstruction module and the spatiotemporal regression module with different color, visualize the temporal representation of different facial patterns and the map of PSD, and highlight the constraints of PSD contrast, coarse-grained reconstruction, and fine-grained rPPG. In addition, to demonstrate our disentanglement logic more intuitively, we represent $\mathbf{F}_{\rm fore}$ (Key, K), $\mathbf{F}_{\rm back}$ (Query, Q), and $\mathbf{F}_{\rm fine}$ (Score) and their attraction and repulsion, respectively.

\section{Experiments}
\label{sec04}

We conduct the experiments based on BUAA-MIHR \cite{xi_fg20}, VIPL-HR \cite{niu_tip20}, MR-NIRP-IND \cite{Nowara_cvpr18}, and -DRV \cite{Nowara_tits22}, which are important public datasets focusing on complex scenes in rPPG. Among them, BUAA-MIHR is divided into 11 light modes with fine span. VIPL-HR is the largest actual indoor dataset, including 3 natural lighting acquisitions and 9 dynamic types. MR-NIRP-IND is a studio dataset with object motion, and its extension MR-NIRP-DRV is the largest outdoor dataset involving real-world driving, extreme illumination, and day-night variations. More details and samples are shown in the Supplementary Materials (Supp).

Since the video frame rate and label frequency of different datasets vary, we uniformly interpolate them to 25 fps. Accordingly, the sliding window length based on BioSE is 100 frames. We annotate facial landmarks based on the face recognition module involved in the publicly available code\footnote{https://github.com/Sachiel0916/repss-track1-top3/}, and build the facial skin STMap based on the first 64 non-overlapping areas ($L$) enclosed by the 68 landmark nodes. On these bases, after spatial scale sampling and unification, the temporal dimension ($T$) of STMap of $\mathbf{S}_{\rm face}$, $\mathbf{S}_{\rm back}$, and $\mathbf{S}_{\rm glob}$ is 320. Our model is based on the PyTorch framework and runs on a device with 4 RTX 4090 GPUs. The optimizer is AdamW, the initial learning rate is 1$\times$10$^{-5}$, 3/4 sample maps in each dataset are randomly utilized as training set, and another 1/4 is tested. They have 100 training epochs and adjust the learning rate to 0.5$\times$10$^{-5}$ after the 50th epoch.

\begin{figure}[t]
	\centering
	\includegraphics[width=0.96\columnwidth]{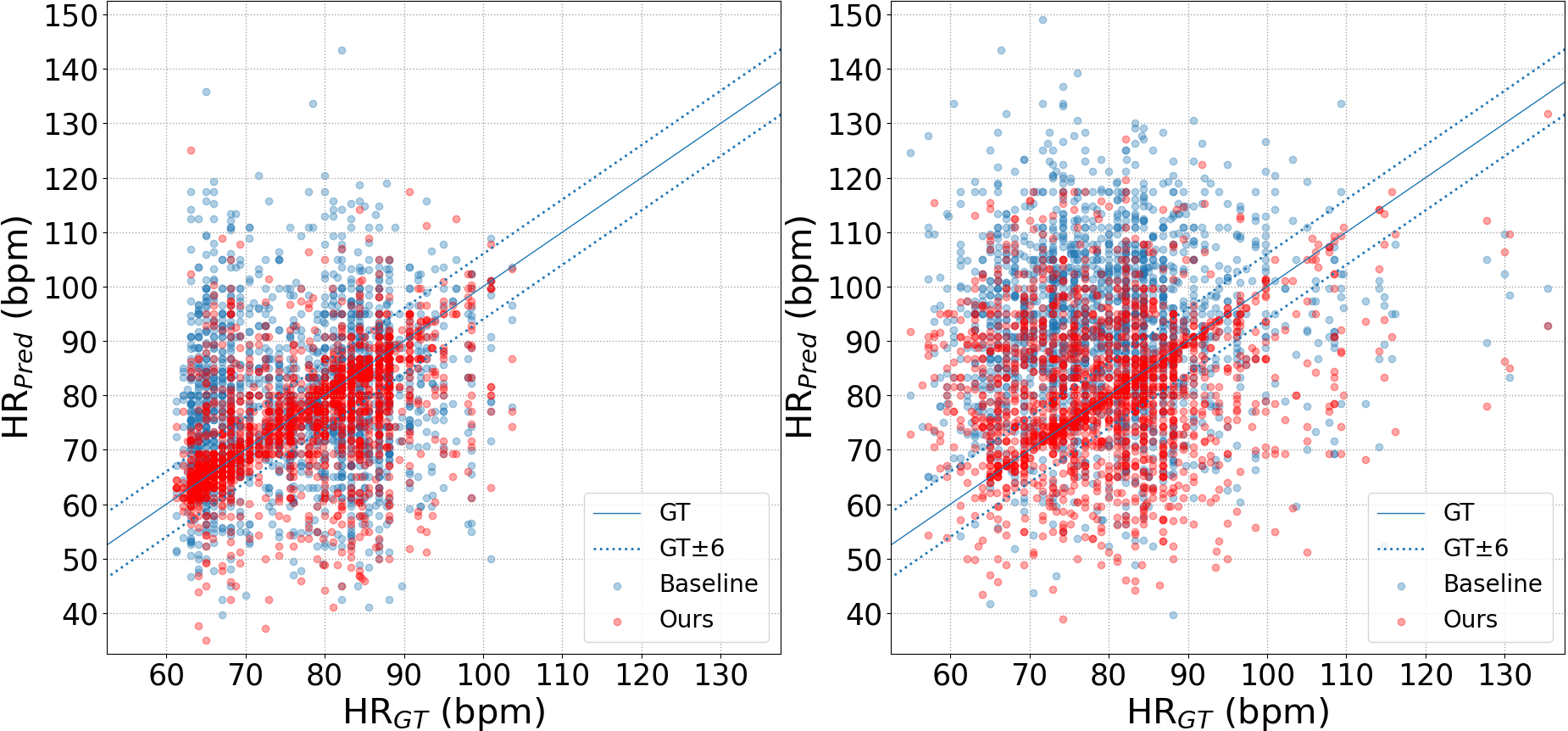}
	\caption{Our method improves upon the state-of-the-art interference disentanglement baseline in both indoor and outdoor scenes.}
	\label{img05}
\end{figure}

\begin{figure}[t]
	\centering
	\includegraphics[width=0.96\columnwidth]{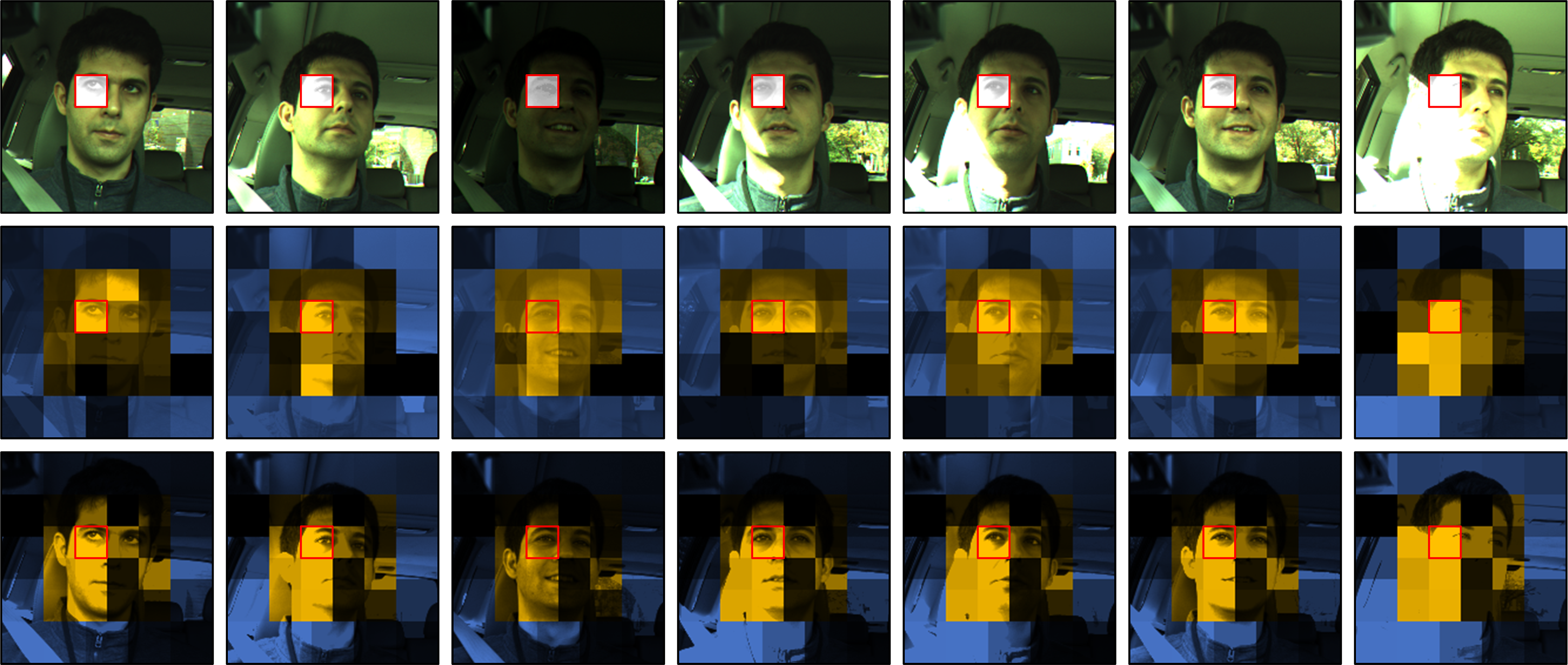}
	\caption{Anchor patch (white) and its dynamic correlations with foreground (yellow) and background (blue) frame sub-blocks.}
	\label{img06}
\end{figure}

\subsection{Comparison with the state-of-the-art methods}
\label{sec41}

We train models independently in BUAA-MIHR and VIPL-HR, and jointly in MR-NIRP-IND and -DRV (MR-NIRP). Afterwards, we validate them across the datasets opposite to each other based on the training weights of MR-NIRP and VIPL-HR. We use the open source rPPG toolbox\footnote{https://github.com/phuselab/pyVHR/} to count and statistically analyze the prediction results, and evaluate them using mean absolute error (MAE), root mean square error (RMSE), and Pearson’s correlation coefficient ($\rho$) indicators. On this basis, we compare our measurements with current representative rPPG methods, including traditional chrominance-based methods \cite{wang_tbme17, haan_tbme13, Pilz_cvpr18}, latest CNN models \cite{liu_nips20, liu_tip24, li_aaai23, lu_cvpr21, chen_eccv18, lu_cvpr23}, and highly competitive transformers \cite{liu_wacv23, yu_ijcv23, zou_24}, which are listed in Tab.~\ref{tab1}. As can be observed that our approach has obvious improvements, especially in MR-NIRP-DRV with extreme time-varying interferences.

\begin{table}[t]
	\begin{center}
		\caption{Computational cost and rPPG performance comparison.}
		\resizebox{\linewidth}{!}{
			\begin{tabular}{l|ccc|c}
				\hline
				{rPPG Methods} & {Parameter} & {FLOPs} & {RTX 4090} & {RMSE}$\downarrow$ \\ \hline
				{TS-CAN \cite{liu_nips20}} & {3.91} M & {110.15} G & {~~~~5.52} ms & {14.59} \\
				{ND-DeeprPPG \cite{liu_tip24}} & {6.05} M & {320.08} G & {~~29.87} ms & {~~7.52} \\
				{PhysFormer++ \cite{yu_ijcv23}} & {9.79} M & {~~49.85} G & {217.07} ms & {~~7.62} \\
				{RhythmFormer \cite{zou_24}} & {3.25} M & {~~38.49} G & {~~29.49} ms & ~~\underline{7.49} \\ \hline				
				{Ours} & {6.03} M & {~~53.04} G & {~~24.64} ms & ~~\textbf{7.09} \\ \hline
		\end{tabular}}
		\label{tab3}
	\end{center}
\end{table}

In-the-wild driving scene is the most challenging aspect of rPPG, and the existing advanced methods \cite{chiu_tim23, Nowara_tits22, comas_icip21, wu_tim22} based on existing MR-NIRP-DRV dataset all need infrared data (e.g., NIR). Based on this, Tab.~\ref{tab2} shows the comparison between our network and theirs, and distinguishes subject and light source interferences such as head motion (Mo.), day driving, night driving, and garage acquisition (Ga.). It can be observed that our model can achieve significant advantages when using only RGB. Meanwhile, since the limited feedback of NIR thermal drive on facial color changes, the effect achieved is mediocre. This also proves that rPPG based on visible light video should be the main focus direction. It is important to note that there is no intersection between Tab.~\ref{tab1} and Tab.~\ref{tab2}. This is because Tab.~\ref{tab1} contains the strongest rPPG models based on natural light, while Tab.~\ref{tab2} is only a specific outdoor study using NIR. We have already verified the outdoor study in Tab.~\ref{tab1} by MR-NIRP-DRV. Furthermore, in Fig.~\ref{img01}, the scene comparison we conducted for methods \cite{liu_tip24, yu_ijcv23, zou_24, li_aaai23, lu_cvpr23} is based on weights trained on MR-NIRP. The scenario (a) corresponds on ideal pattern, and (b) matches the combination of daytime and night with motion.

\subsection{Visualization and effectiveness analysis}

To show our model's efforts in real-world interference filtering, we compare it with the most valuable noise disentangl-ement design: ND-DeeprPPG \cite{liu_tip24}. We train corresponding models based on MR-NIRP and draw HR estimation results for 2,000 random periods as scatters. Fig.~\ref{img05} left plot is the garage results, and its right plot is the outdoor HRs following Sec.~\ref{sec41}. It can be clearly found that compared with the adversarial-based baseline, the paradigm we proposed can make predicted values more consistent with GT. It can make more points fall within confidence interval areas of GT$\pm$6 bpm, as well as its achievements in overcoming overfitting at marginal HR points. Comparative scatters with the latest transformer-based rPPGs \cite{yu_ijcv23, zou_24} are shown in the Supp.

To intuitively demonstrate that our learned features can effectively avoid interference, we select a set of video segment from MR-NIRP-DRV and display them in Fig.~\ref{img06}. We use the white patch in the first row as the query anchor. The second row is the spatial similarity between the anchor and each sub-block in the foreground (yellow) and background (blue) of a single frame. The greater the similarity, the more saturated the color. The third row is the temporal correlation between captured sequence frames. For ease of visualization, we utilize window blocks and 4$\times$4 frame scale instead of node STMap. It can be seen that the learned inter-frame correlation obviously tends to the overall temporal distribution. Therefore, for the drastic interference caused by time-varying illumination, we can still stably mine the temporal correlation between different sub-blocks and get rid of the limitations of spatial texture and color. In addition, we draw the actual predicted BVP signals and HR values in the Supp.

\begin{table}[t]
	\begin{center}
		\caption{Comparison of the latest rPPG towards disentanglement.}
		\label{tab4}
		\resizebox{\linewidth}{!}{
			\begin{tabular}{l|c|ccc|ccc}
				\hline
				\multirow{2}{*}{Models} & \multirow{2}{*}{1$\times$} & \multicolumn{3}{c|}{Temporal Sampling} & \multicolumn{3}{c}{Temporal Frequency} \\ \cline{3-8}
				\multirow{2}{*}{} & \multirow{2}{*}{} & {80} & {160} & {640} & {2$\times$} & {4$\times$} & {8$\times$} \\ \hline
				{DeeprPPG \cite{liu_fg20}} & {11.27} & {12.77} & {10.46} & {14.90} & {13.20} & {15.14} & {17.39} \\
				{ND-DeeprPPG \cite{liu_tip24}} & ~~\underline{6.54} & ~~\underline{8.07} & ~~\textbf{6.76} & ~~\underline{9.09} & ~~\underline{7.66} & ~~\underline{8.78} & \underline{10.09} \\ \hline
				{Ours w/o ND} & ~~{8.12} & ~~{9.50} & ~~{8.99} & ~~{9.80} & ~~{9.29} & {11.39} & {13.09} \\
				{Ours} & ~~\textbf{6.03} & ~~\textbf{7.16} & ~~\underline{6.80} & ~~\textbf{7.46} & ~~\textbf{6.68} & ~~\textbf{8.09} & ~~\textbf{9.12} \\ \hline
		\end{tabular}}
	\end{center}
\end{table}

\begin{figure}[t]
	\centering
	\includegraphics[width=1\columnwidth]{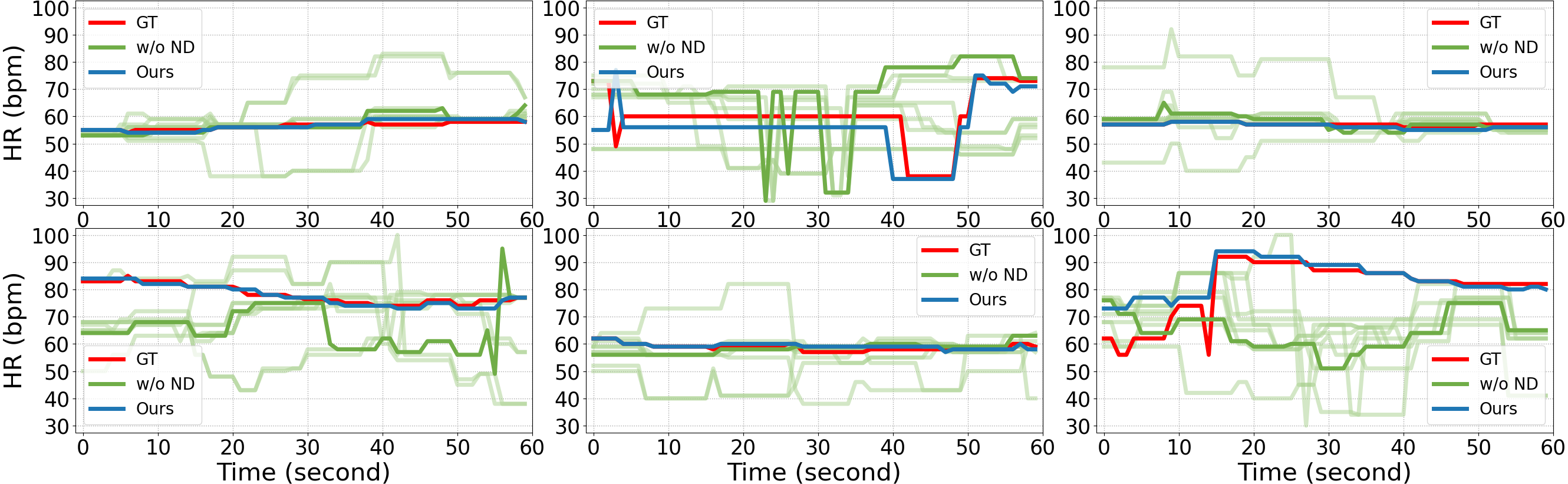}
	\caption{Comparison with and without disentanglement module.}
	\label{img07}
\end{figure}

Tab.~\ref{tab3} presents the advantage of our network in computational complexity. Letting RMSE results under VIPL-HR as example evidence, due to our lightweight layout, we can achieve the best performance without significantly increasing the number of parameters and floating point operations per second (FLOPs), even lower than the cost of the latest rPPG methods. More comparisons are shown in the Supp.

\subsection{Ablation studies}

To investigate the effect of the core component of our architecture, the interference disentanglement module. We compare our whole model with the learning results of VIPL-HR without the disentanglement mechanism (w/o ND), and the continuous one-minute HR estimation results are plotted in Fig.~\ref{img07}. At this point, to simultaneously discuss the impact of the time window length on the prediction, we set the STMap time dimension of the undisentangled input size to 300, 280, 260, 240, 220, 200, 180, 160, and 140 (corresponding light-colored lines). As can be seen that although increasing the temporal length can improve the model's fit to BVP and thus calculate HR values, it still performs poorly. In addition, we present more exhibits and scatters in the Supp, including the ablation studies on the disentanglement module and BioSE.

Based on the above studies on unwrapping and window, we further compare with ND-DeeprPPG \cite{liu_tip24} in MR-NIRP-DRV. Similar to it with or without ND module (DeeprPPG \cite{liu_fg20}), our configuration follows: We apply varying time dimensions (80, 160, 320, 640 frames) and readjust frequencies (2$\times$, 4$\times$, 8$\times$) of STMap as training and testing samples. Specifically, temporal sampling is to training the models with different video clip lengths, and temporal frequency refers to testing the corresponding sets at different frequencies. RMSE results are listed in Tab.~\ref{tab4}, which illustrates the reasons for our choice of window and the advantages of our manner over existing rPPG disentanglement strategies.

\begin{figure}[t]
	\centering
	\includegraphics[width=0.96\columnwidth]{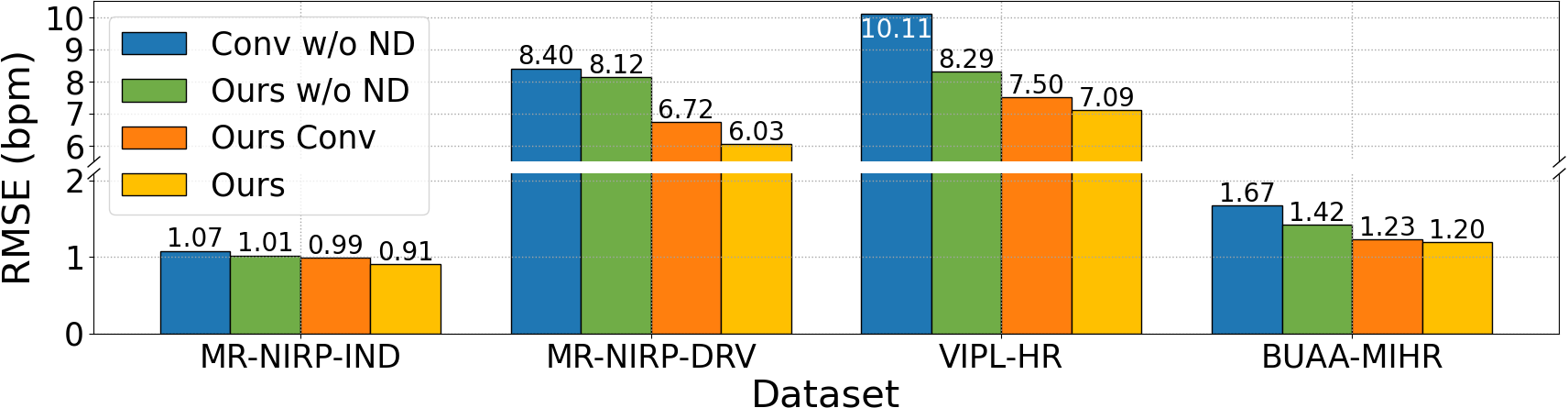}
	\caption{Results of four different structures on validation datasets.}
	\label{img08}
\end{figure}

While building the framework based on transformer, we discuss the perception of our disentanglement in CNN. This fully verifies that the improvement comes from architecture design rather than the inherent self-attention of transformer. We show the effects on four datasets based on CNN (Conv) and transformer. RMSE results in Fig.~\ref{img08} can be adopted as a test of the rationality of our design. In addition, we compare the efficiency of CNN based on our strategy in the Supp.

\begin{figure}[t]
	\centering
	\includegraphics[width=0.96\columnwidth]{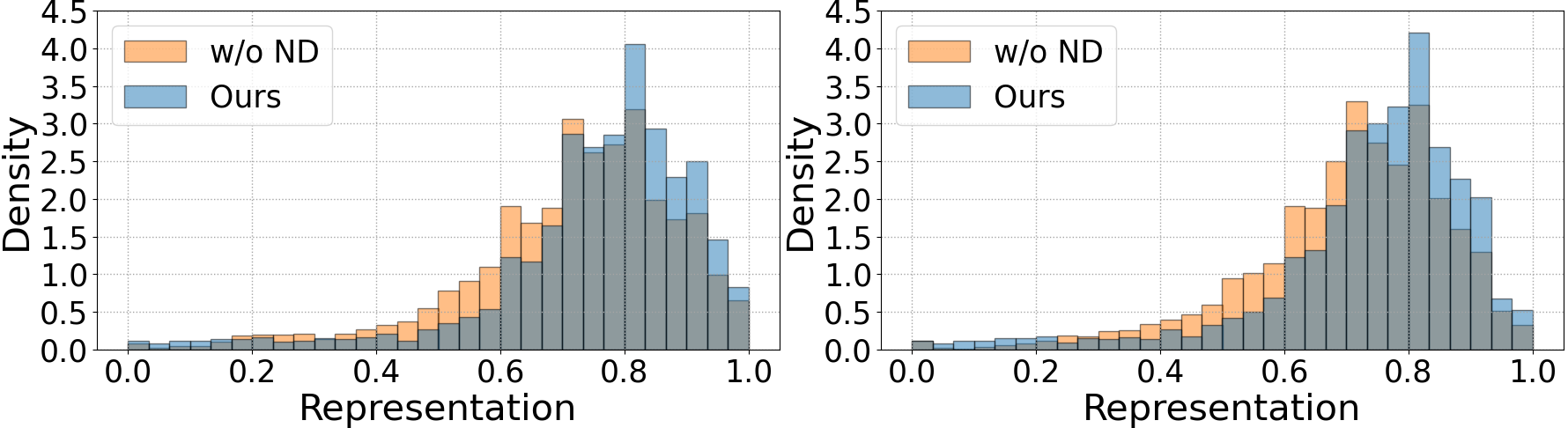}
	\caption{Representation of the learned map corresponding to BVP peaks for daytime and night real-world driving scenarios.}
	\label{img09}
\end{figure}

Finally, for the pure outdoor environments and complex time-varying lighting, we calculate the representation of the learned features corresponding to the pulsing spike points on the disentangled and undisentangled maps of MR-NIRP-DRV in day (left) and night (right) driving scenes, and count their normalized representations. Fig.~\ref{img09} shows that the peak distribution learned by our overall structure is closer to 1.0, which means that its sensing has a higher overlap with GT.

\section{Conclusion}

Focusing on remote HR prediction and analysis has a wide range of practical applications. This paper proposes a novel rPPG framework to overcome the complex environment and eliminate the extreme interference that are difficult to solve with current approaches. Benefiting from our improved spatiotemporal map reconstruction, self-supervised contrastive learning, and temporal regression, we can robustly perform contactless measurement and achieve superior effects, especially in outdoor scenarios. Besides, we not only verify it on visible data, but also study infrared images, and prove that optical reflectance has significant roles over thermal drive.

\section{Acknowledgments}

\small{This work was supported in part by the National Natural Science Foundation of China under Grants: 62361166670, U24A20330, 62176124, and 62276135.}

\small
\bibliographystyle{ieeenat_fullname}


\end{document}